\documentclass[10pt, a4paper]{article}

\usepackage{lrec2026}

\usepackage{booktabs}
\usepackage{amsmath}
\usepackage{pifont}
\usepackage{multirow}
\usepackage{array}
\usepackage{enumitem}
\usepackage{todonotes}
\usepackage{blindtext}
\usepackage{ulem}
\usepackage{xcolor}
\usepackage{wrapfig}
\newcolumntype{H}{>{\setbox0=\hbox\bgroup}c<{\egroup}@{\hspace*{-\tabcolsep}}}
\newcommand{\dataset}{\texttt{INDI-PROP}}
\newcommand{\modelNarrative}{\texttt{FANTA}}
\newcommand{\modelTechnique}{\texttt{TPTC}}
\definecolor{TPColor}{HTML}{28A745} 
\definecolor{FPColor}{HTML}{DC3545} 
\definecolor{FNColor}{HTML}{E8A900} 

\definecolor{darkgreen}{rgb}{0,0.5,0}

\usepackage{tcolorbox}
\tcbuselibrary{skins, breakable}

\usepackage{tcolorbox}

\tcbset{
  colback=gray!5!white,
  colframe=black!70,
  fonttitle=\bfseries,
  boxrule=0.6pt,
  arc=2pt,
  outer arc=2pt,
  left=4pt,right=4pt,top=4pt,bottom=4pt,
  enhanced,
  breakable,
}

\usepackage{graphicx}
\usepackage{subcaption}
\usepackage{xcolor} 
\usepackage{array} 
\usepackage{graphicx}
\usepackage{booktabs}
\usepackage{xcolor} 
\usepackage{array}  

\title{Fine-grained Narrative Classification in Biased News Articles}

\name{
\normalsize{
\textbf{Zeba Afroz}$^{1}$,
\textbf{Harsh Vardhan}$^{1}$,
\textbf{Pawan Bhakuni}$^{2}$, }\\
\normalsize{\textbf{Aanchal Punia}$^{2}$,
\textbf{Rajdeep Kumar}$^{2}$,
\textbf{Md.\ Shad Akhtar}$^{1}$
}
}

\address{
$^{1}$Indraprastha Institute of Information Technology Delhi\\
$^{2}$Bharat Electronics Limited\\
\texttt{zebaa@iiitd.ac.in, harsh25001@iiitd.ac.in, shad.akhtar@iiitd.ac.in}\\
\texttt{pwn.bhakuni@gmail.com, aanchal.nith@gmail.com, rajdeep8709@gmail.com}
}

\abstract{
Narratives are the cognitive and emotional scaffolds of propaganda. They organize isolated persuasive techniques into coherent stories that justify actions, attribute blame, and evoke identification with ideological camps. In this paper, we propose a novel fine-grained narrative classification in biased news articles. We also explore article-bias classification as the pre-cursor task to narrative classification. We develop \dataset, the first ideologically grounded fine-grain narrative dataset with multi-level annotation for analyzing propaganda in Indian news media. Our dataset \dataset{} comprises 1,266 articles focusing on two polarizing socio-political events in recent times: CAA and the Farmers’ protest. Each article is annotated at two hierarchical levels: (i) ideological article-bias (\textit{pro-government}, \textit{pro-opposition}, \textit{neutral}) and (ii) event-specific fine-grained narrative frames anchored in ideological polarity and communicative intent.
We propose \modelNarrative\ a GPT-4o-mini guided multi-hop prompt-based reasoning framework for bias and narrative classification. \modelNarrative\ leverages multi-layered communicative phenomenon by integrating information extraction and contextual framing for hierarchical reasoning.  
}

\begin{document}

\maketitleabstract

\section{Introduction}

Propaganda refers to the systematic use of language, images, and framing to shape public perception, influence opinion, and advance ideological agendas. The biased connotation operates through selective presentation, emotional manipulation, and repetition to reinforce obscure objectivity and polarize discourse. 

By distorting factual reporting and embedding ideological bias in everyday news consumption, it can erode trust in institutions, normalize misinformation, and reinforce echo chambers. Therefore, tackling it is not only a question of fact-checking but also of understanding how language constructs different versions of reality. 

Detecting propaganda at scale is central to ensuring media transparency, civic literacy, and informed decision-making. Early research in computational propaganda focused on identifying persuasive techniques that constitute the linguistic building blocks of manipulative discourse. This line of work was systematized through several shared tasks, including a series of SemEval workshops during 2020 -- 2024 \cite{da-san-martino-etal-2020-semeval, dimitrov-etal-2021-semeval, piskorski-etal-2023-semeval, dimitrov-etal-2024-semeval}. The scope of these work varies from detecting persuasive techniques at the span level to incorporation of multimodality through memes. 
Although these datasets advanced the understanding of how propaganda operates linguistically, they remain predominantly in English \cite{maarouf2023hqp, gruppi2021nela} and Arabic \cite{al2025multiprose, alam-etal-2022-overview}, overlooking the socio-political and cultural context of countries like India. Moreover, these effort primarily explore surface layer of persuasion to identify how manipulation is expressed without revealing what storyline, views, or narratives it promotes. 

Narratives are the cognitive and emotional scaffolds of propaganda. They organize isolated persuasive techniques into coherent stories that justify actions, attribute blame, and evoke identification with ideological camps. Understanding narratives enables us to go beyond token-level detection and reveal how entire discourse ecosystems construct different realities. By modeling narrative explicitly, we can trace the mechanism of polarization, measure ideological symmetry between opposing sides, and build systems capable of narrative-aware propaganda analysis. Recently, SemEval-2025 Task 10 \cite{piskorski-etal-2025-semeval} studied narratives in propaganda in a global setting. However, these taxonomies treat narratives as thematic clusters without embedding ideological bias of the article.

\begin{figure}[!t]
\centering
\includegraphics[width=0.8\columnwidth]{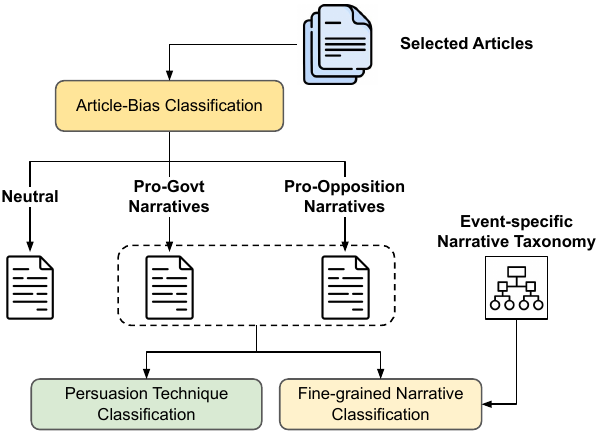}
\caption{Multi-level schema of \dataset.}
\label{fig:problem}
\vspace{-5mm}
\end{figure}

To address the problem, we propose a multi-stage interpretive task to identify underlying storyline that links individual persuasive acts into a coherent ideological message. In particular, we explore three subtasks within the propaganda framework: a) \textbf{article-bias classification} to identify the ideological persuasion as \textit{pro-government}, \textit{pro-opposition}, or \textit{neutral}; b) a novel \textbf{taxonomy-driven narrative classification} to reveal the underlying storyline; and c) \textbf{persuasive technique classification} used for propagating the agenda.     

To support the tasks, we develop an ideologically grounded dataset, \dataset, for two recent socio-political events in India: Farmers' protest and CAA. Inspired by the hierarchical narrative taxonomy of \newcite{dash2022narrative}, for each event, we propose 2-level deep taxonomy to cover pro-government and pro-opposition aligned narratives. Each narrative in our framework is explicitly anchored to an ideological side, a communicative intent (legitimization, vilification, glorification, etc.), and a discursive focus that reveals how one side represents events, institutions, or social groups. In total, \dataset{} comprises of 1266 articles with three bias labels, 20 narrative labels across two events, and 20 persuasion techniques. 

We also propose two frameworks employing multi-hop prompt-based reasoning for the three classification tasks. We propose \modelNarrative{} (\textbf{F}ine-gr\textbf{A}ined \textbf{N}arra\textbf{T}ive Cl\textbf{A}ssification) that leverages \textit{entity-relation dynamics} and \textit{context framing} for the article-bias and fine-grained narrative classification. In persuasive technique classification, conventionally, fine-tuned PLMs usually outperforms LLMs \cite{jose2025large, hasanain-etal-2024-gpt}. However, we argue that due to subtle persuasive language LLMs struggles to appropriately decompose the article for underlying persuasive techniques. To counter this challenge, we propose \modelTechnique{}, a novel \textbf{T}wo-stage \textbf{P}ersuasion \textbf{T}echnique \textbf{C}lassification that leverages the coarse persuasive intent before mapping it to fine-grained techniques. We evaluate both frameworks on \dataset{} using GPT-4o-mini as the grounding LLM and observe performance-gain against fine-tuned PLMs.

\paragraph{Contributions:} We summarize contributions as:
\begin{itemize}[leftmargin=*, nolistsep, noitemsep]
    \item We propose a novel task of fine-grained narrative classification to provide interpretability of the underlying stories.
    \item We present \dataset, a multi-level ideologically aligned dataset with annotations for bias, narratives, and persuasive techniques.
    \item We benchmark \dataset\ using a multi-hop prompt-based inference only model, \modelNarrative{}, for the fine-grained narrative classification and a novel two-stage persuasive intent decomposition model, \modelTechnique{}, for the technique identification.
\end{itemize}

\begin{table*}[t]
\centering

\resizebox{\textwidth}{!}{
\begin{tabular}{p{1.5cm} p{12cm} l l l}
\toprule
{\bf Event} & {\bf Article}  & {\bf Bias} & {\bf Technique} & {\bf Narrative} \\ 

\midrule
CAA & \multirow[t]{2}{12cm}{The reality is that after a detailed analysis of the provisions of the Act, we found that the law has nothing to do with Indian Muslims", the Chief added....} & Pro-Govt & {Assertion} & Glorification of CAA \\
& & & {Glittering Generalities (Virtue)} & Opposition spreading misinformation and fear \\

\midrule
CAA & \multirow{2}{12cm}{Across the country, as protests against the contentious Citizenship (Amendment) Act (CAA), NRC and police violence on students escalate, ....} & Pro-Opp & {Loaded Language} & Glorifying Anti-CAA Protesters
 \\

& & & {Flag Waving}  & Framing Anti-CAA Protesters as Victims \\


\midrule
\multirow[t]{3}{1.5cm}{Farmer's Protest} & \multirow{3}{12cm}{The list has doctors, lawyers, sportspersons, farmers, youth, women, and former IAS," said Bharatiya Janata Party (BJP) General Secretary Tarun Chugh Amid hectic campaigning ....} & Pro-Govt & {Smears} & Glorification of Central Government
 \\
& & & & Vilification of Opposition \\
\\


\midrule
\multirow{2}{1.5cm}{Farmer's Protest} & \multirow[t]{2}{12cm}{Amid growing outrage over the Lakhimpur Kheri violence, Rahul Gandhi questioned PM Modi's silence on the alleged murders of farmers, .....} & Pro-Opp & {Name Calling \& labelling}  & Vilification of Central government   \\
& & & {Red Hearing} & Depicting Farmers as Victims \\
\bottomrule
\end{tabular}}
\vspace{-3mm}
\caption{A few annotated samples from \dataset{}.}
\label{tab:dataset_sample}
\vspace{-3mm}
\end{table*}


\section{Related Work}
Most of the prior research on propaganda are focused on linguistically identifying persuasive techniques. \newcite{da-san-martino-etal-2019-fine, da-san-martino-etal-2020-semeval} introduced the first large-scale benchmark for detecting 18 persuasive techniques in English news articles. Subsequent shared tasks in SemEval (2021 and 2024) \cite{dimitrov-etal-2021-semeval, dimitrov-etal-2024-semeval} extended it to multimodality, incorporating memes as visual elements. In SemEval-2023, \newcite{piskorski-etal-2023-semeval} further extended it to multilinguality. These datasets have been instrumental in modeling propaganda linguistically.
However, these resources remain predominantly western-centric, reflecting US \& European media ideologies, whereas, non-Western democracies, like India, with its diverse ideological and complex politics, remained underrepresented. Indian efforts have primarily focused on fact-checking \cite{singhal2022factdrill} and hate-speech detection \cite{k-b-2025-ssncse}, with limited exploration in the interplay between persuasion, ideology, and narrative framing. Notably, \newcite{dash2022narrative} proposed a hierarchical narrative tree structure over Indian Twitter data, offering early insights into narrative construction in political discourse.


A complementary line of research has explored framing and narrative analysis in political communication. The Policy Frames Codebook \cite{boydstun2014tracking}  and the Media Frames Corpus \cite{card2015media} examined how policy issues are framed. More recent studies, SemEval-2025 Task 10 \cite{piskorski-etal-2025-semeval}, formalized the multilingual characterization of narratives. Yet, these works typically define narratives as thematic clusters without embedding ideological polarity. Methodologically, previous work has employed transformer-based BERT \cite{devlin2019bert}, RoBERTa \cite{liu2019roberta}, and DeBERTa \cite{he2020deberta}, etc., for persuasive technique detection, which perform well in token-level tasks. However, LLMs despite their strong linguistic understanding, struggle to detect persuasive techniques in zero or few-shot settings \cite{jose2025large, hasanain-etal-2024-gpt}, as persuasion often depends on subtle, context-driven cues beyond single-step inference.

\paragraph{Novelty in our work:} While prior research has studied bias detection and persuasive technique identification, our principal contribution lies in the new task of fine-grained narrative classification. Moreover, unlike prior corpora, \dataset{} presents ideologically grounded fine-grained narratives along with persuasive techniques and bias annotations. This joint representation enables the analysis of how persuasive cues accumulate into broader ideological storylines. Furthermore, our approach of decomposing the narrative classification and persuasive techniques identification tasks into a multi-hop strategy inspires LLMs to comprehend the conceptual category of persuasion in an efficient manner, against state-of-the-art fine-tuned PLMs in existing works.

\section{Dataset}

We develop \dataset, an ideologically grounded fine-grained narrative dataset that includes annotations of article-bias labels, taxonomy-driven narrative labels, and 20 persuasive techniques. The dataset centers on two major socio-political events in India, i.e., Citizenship Amendment Act (CAA) and the Farmers' protest. As shown in Fig \ref{fig:problem}, each article undergoes multi-level annotation schema for covering bias, fine-grained narrative, and persuasive techniques for both events. This setup enables us to capture how persuasive language, ideological bias, and narrative framing co-occur and interact within Indian news media. A few examples of the \dataset{} dataset is depicted in Table \ref{tab:dataset_sample}.

\subsection{Data Collection}

We select events based on following criteria: (a) it must span a prolonged duration, allowing us to trace how narratives emerge and evolve; (b) it must be highly polarizing, ensuring the presence of clear ideological divisions and contrasting perspectives; and (c) have sustained public engagement which have the potential to influence everyday discourse rather than being isolated incidents. Based on these criteria, we select two major socio-political events in India in recent times, i.e., the Citizenship Amendment Act/National Register of Citizens (CAA/NRC) protests (2019 to 2024) and the farmers' protest (2020 to 2024). Both are long running, mass mobilized, and ideologically charged movements offering media discourse for studying narrative construction.

For data sources, we refer to Media Bias/Fact Check (MBFC)\footnote{\label{fn:mbfc}https://mediabiasfactcheck.com/}, to identify outlets that represent an ideological spectrum. We include OpIndia, Republic World, Swarajya, The Quint, and Hindustan Times, covering right to left leaning perspectives. We collect articles using BeautifulSoup through a keyword driven web scrapping, employing terms such as \textit{CAA}, \textit{NRC}, \textit{Farm Laws}, \textit{Farmers' protest}, \textit{Kisan Andolan}, etc. 

\paragraph{Data Cleaning:} We perform a systematic data cleaning process to ensure the quality and relevance of the collected corpus. We first remove duplicate entries by comparing URLs, titles, and content similarity. Next, we filter out of scope articles published outside the defined timeframes.

\begin{figure*}[!t]
    \centering
    \begin{subfigure}[b]{0.45\textwidth}
        \centering
        \includegraphics[width=\textwidth]{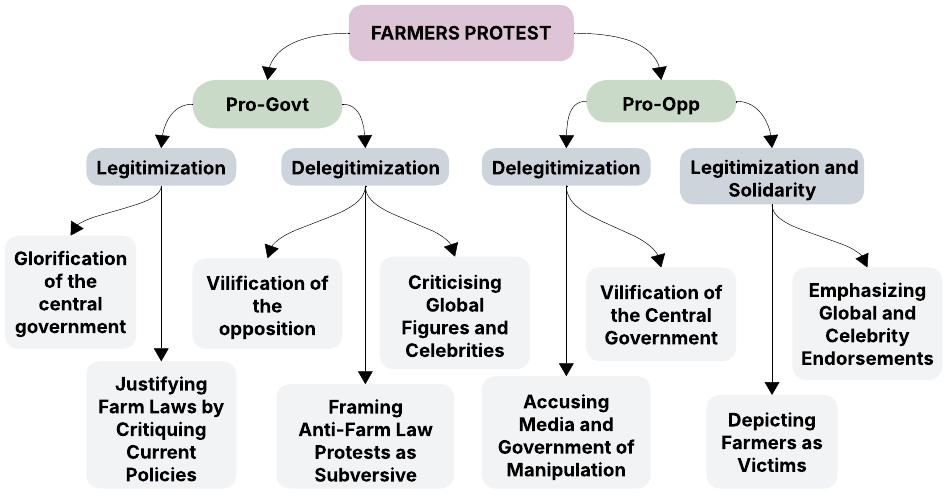}
        \caption{Narrative Taxonomy for Farmer's Protest}
        \label{fig:taxonomy:farmer}
    \end{subfigure}
    \begin{subfigure}[b]{0.45\textwidth}
        \centering
        \includegraphics[width=\textwidth]{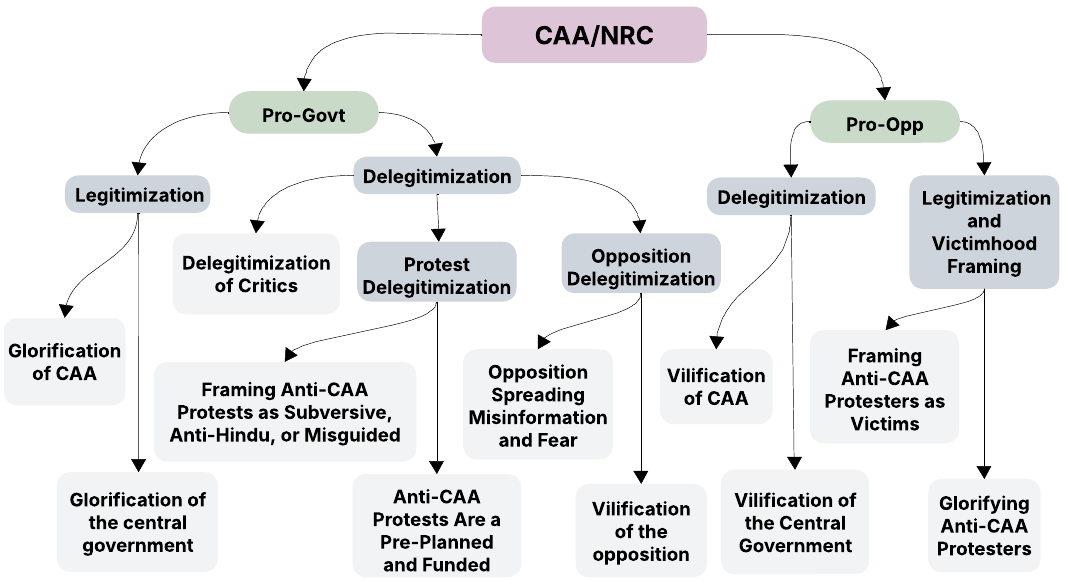}
        \caption{Narrative Taxonomy for CAA/NRC}
        \label{fig:taxonomy:CAA}
    \end{subfigure}
    \vspace{-3mm}
    \caption{Narrative Taxonomy for two socio-political movements: Farmers' Protest and CAA/NRC. Taxonomy is organized along the ideological orientation of Pro-Govt and Pro-Opp.}
    \label{fig:taxonomy}
    \vspace{-3mm}
\end{figure*}

\subsection{Annotation Guidelines}

The objective of the annotation process was to construct a qualitative resource that captures article-level ideologically grounded fine-grained narrative framing along with persuasive techniques within Indian spectrum.

\paragraph{Article-Bias Guidelines:} Bias refers to the systematic ideological inclination. The task is to determine the overall ideological leaning of an article, whether it aligns with \textit{pro-government} (Right) or\textit{ pro-opposition} (Left), or \textit{neutral} reporting:

\begin{itemize}[leftmargin=*, noitemsep, nolistsep]
    \item \textbf{Pro-Govt:} Article aligns with pro-government, nationalist, or conservative narratives.\\
     \begin{small}{E.g.: \textit{"The government's reform has empowered farmers to sell freely without a middleman" → Pro-Govt}} \end{small}
     
    \item \textbf{Pro-Opp:} Article aligns with pro-opposition, liberal, or activist perspectives that question or critique government action.\\
    \begin{small}{E.g.: \textit{"The new farm laws are a betrayal of the farmer's trust and threaten livelihoods" → Pro-Opp}} \end{small}
    
    \item \textbf{Neutral: } Article presents balanced reporting or lacks evaluative language. \\
    \begin{small}{E.g.: \textit{"The Parliament passed three new agricultural bills on Monday" → Neutral}} \end{small}
\end{itemize}

\noindent Annotators were instructed to read the entire article and consider tone, framing, evaluative language, and attribution of responsibility before assigning the labels. 
 
\paragraph{Fine-Grain Narrative Guidelines:} After assigning the article-level bias label, annotators proceed to identify the dominant narrative expressed in each article. A narrative represents the storyline through which propaganda constructs legitimacy, blame, and moral position, and is unique for each event. It connects surface-level persuasive cues to deeper ideological framing, illustrating how each side legitimizes itself and delegitimizes its opponent. We utilize event-specific narrative taxonomies for fine-grained narrative classification. 

\textit{\ul{Narrative taxonomy}:}\label{sec:taxo}
We create separate narrative taxonomies for both events as depicted in Figure~\ref{fig:taxonomy}. We carefully analyze each event to understand its timeline and how it unfolded during its course. Each taxonomy was constructed after a detailed review of media coverage timelines, political speeches, protest statements, and counter narratives, to ensure a comprehensive mapping of the ideological space. The narratives for both events follow a hierarchical taxonomy that organizes ideological framing from broad to specific. The root node represents the overarching socio-political event. The second level divides the discourse into primary ideological clusters: \textit{pro-government} (right or conservative) and \textit{pro-opposition} (left or liberal). Subsequent branches captures broader narrative themes unique to each ideological side. 

\noindent Following instructions are given to annotators:

\begin{itemize}[leftmargin=*, noitemsep, nolistsep]
    \item Narrative labels are applicable only for Pro-Govt and Pro-Opp biased articles.
    
    \item Refer to the event-specific taxonomy 
    and select the leaf node whose description best captures the overall framing of the article. The chosen narrative should reflect the article's dominant ideological storyline rather than isolated sentences or quotations. In particular, annotators should identify:
    \begin{itemize}[leftmargin=*, noitemsep,nolistsep]
        \item \underline{Legitimization:} Article portrays a particular side as just, necessary, patriotic, or reformative: analyze the article for evaluative language, success framing, and moral justification.
        \item \underline{Delegitimization:} Article undermines or attacks the credibility of the other party using blame, ridicule, misinformation, or moral discredit: analyze the article for lexical cues implying corruption, chaos, betrayal, or threat.
        \item Based on above dominant framing, annotate the fine-grained narrative using the event-specific taxonomy.
    \end{itemize}
\end{itemize}

\begin{table*}[h!]
\centering

\subfloat[CAA/NRC\label{tab:stat:narrative:caa}]{
\resizebox{0.49\textwidth}{!}
{
\begin{tabular}{l l rrr}
\toprule
\textbf{Stance} & \textbf{Narrative class} & Train & Test & Total  \\
\midrule

\multirow{7}{*}{\textbf{Pro-Govt}} &	\textbf{C1}: Glorification of the central government	&	100	&	27	&	127	\\
&	\textbf{C2}: Vilification of the opposition	&	155	&	48	&	203	\\
&	\textbf{C3}: Glorification of CAA	&	59	&	19	&	78	\\
&	\textbf{C4}: Delegitimization of Critics	&	155	&	42	&	197	\\
&	\textbf{C5}: Framing anti-CAA protests as Subversive, Anti-Hindu, or Misguided	&	122	&	36	&	158	\\
&	\textbf{C6}: Opposition spreading misinformation and fear	&	119	&	30	&	149	\\

&	\textbf{C7}: Anti-CAA protests are a pre-planned and funded conspiracy	&	49	&	11	&	60	\\ \midrule
\multirow{4}{*}{\textbf{Pro-Opp}} &	\textbf{C8}: Vilification of the central government	&	80	&	13	&	93	\\
&	\textbf{C9}: Vilification of CAA	&	68	&	9	&	77	\\
&	\textbf{C10}: Glorifying anti-CAA protesters	&	57	&	6	&	63	\\
&	\textbf{C11}: Framing anti-CAA protesters as victims	&	50	&	3	&	53	\\
\midrule
&   Total & 1014 & 244 & 1258 \\
\bottomrule
\end{tabular}
}}
\hfill
\subfloat[Farmer's Protest\label{tab:stat:narrative:farmer}]{
\resizebox{0.47\textwidth}{!}{
\begin{tabular}{l l rrr}
\\
\toprule
\textbf{Stance} & \textbf{Narrative class} & Train & Test & Total  \\
\midrule
\multirow{4}{*}{\textbf{Pro-Govt}} &	\textbf{F1}: Glorification of the central government	&	15	&	6	&	21	\\
&	\textbf{F2}: Vilification of the opposition	&	19	&	7	&	26	\\
&	\textbf{F3}: Justifying farm laws by critiquing current policies	&	15	&	5	&	20	\\
&	\textbf{F4}: Criticizing global figures and celebrities	&	2	&	2	&	4	\\ \midrule
\multirow{5}{*}{\textbf{Pro-Opp}} &	\textbf{F5}: Vilification of the central government	&	115	&	19	&	134	\\
&	\textbf{F6}: Depicting farmers as victims	&	117	&	20	&	137	\\
&	\textbf{F7}: Framing anti-farm law protests as subversive	&	12	&	6	&	18	\\
&	\textbf{F8}: Accusing media and government of manipulation	&	9	&	1	&	10	\\
&	\textbf{F9}: Emphasizing global and celebrity endorsements	&	6	&	1	&	7	\\
\midrule
&   Total & 310 & 67 & 377 \\
\bottomrule
\end{tabular}
}}
\vspace{-3mm}

\caption{Dataset statistics for the fine-grained narrative classes as per the taxonomy.}
\label{tab:stat_narrative}
\vspace{-5mm}
\end{table*}

\paragraph{Pro-Govt Aligned Narratives (Farmer’s protest):}

\begin{enumerate}[leftmargin=*, noitemsep, nolistsep]
\item \textbf{Glorification of the Central Government:} Articles portraying the central government as visionary, reformist, or acting in farmers’ best interest.
\begin{itemize}[leftmargin=*, noitemsep, nolistsep]
    \item Presents farm laws as historic, bold, or progressive.
    \item Credits leadership for modernization or economic reform.
    \item Highlights efficiency, decisiveness, or long-term vision.
    \item \texttt{Eg:} \textit{"The PM's reform vision will finally free farmers from middlemen.", "These laws mark a new era of prosperity for rural India."}
\end{itemize}

\item \textbf{Justifying Farm Laws by Critiquing Existing Policies:} Articles defending new farm laws by emphasizing flaws or inefficiencies of the older system.
\begin{itemize}[leftmargin=*, noitemsep, nolistsep]
    \item Frames previous market structures as exploitative or outdated.
    \item Argues that reforms are necessary corrections.
    \item \texttt{Eg:} \textit{"For decades, farmers were trapped in the APMC monopoly.", "The old system benefited brokers, not farmers."}
\end{itemize}

\item \textbf{Vilification of the Opposition:} Articles blaming opposition parties or leaders for politicizing or misleading farmers.
\begin{itemize}[leftmargin=*, noitemsep, nolistsep]
    \item Claims protests are driven by opposition propaganda.
    \item Describes leaders as hypocritical or self-serving.
    \item \texttt{Eg:} \textit{"Opposition leaders are inciting unrest for political mileage.", "Those who failed farmers for 60 years now pretend to care."}
\end{itemize}

\item \textbf{Framing Anti-Farm Law protests as Subversive:} Articles depicting the protests as anti-national or disruptive to peace.
\begin{itemize}[leftmargin=*, noitemsep, nolistsep]
    \item References funding or conspiracy.
    \item Describes protests as violent, separatist, or extremist.
    \item \texttt{Eg:} \textit{"These so-called protests are an attempt to destabilize the nation."}
\end{itemize}

\item \textbf{Criticizing Global Figures and Celebrities:} Articles discrediting international figures or celebrities who expressed solidarity with farmers.
\begin{itemize}[leftmargin=*, noitemsep, nolistsep]
    \item Dismisses foreign commentary as ignorant or biased.
    \item Emphasizes domestic sovereignty over external opinions.
    \item \texttt{Eg:} \textit{"Foreign celebrities should stop lecturing India without understanding facts.", "These global influencers are tools of anti-India campaigns."}
\end{itemize}
\end{enumerate}

\paragraph{Pro-Opp Aligned Narratives (Farmer’s protest):} 

\begin{enumerate}[leftmargin=*, noitemsep, nolistsep]
\item \textbf{Vilification of the Central Government:} Articles accusing the government of arrogance, repression, or lack of empathy toward farmers.
\begin{itemize}[leftmargin=*, noitemsep, nolistsep]
    \item Frames government as authoritarian or indifferent.
    \item Uses moral condemnation or emotional tone.
    \item \texttt{Eg:} \textit{"The government silences farmers instead of hearing their pain.", "Farmers' pleas are ignored by those in power."}
\end{itemize}

\item \textbf{Depicting Farmers as Victims:} Articles portraying farmers as suffering under unjust laws or state apathy.
\begin{itemize}[leftmargin=*, noitemsep, nolistsep]
    \item Focuses on hardships, deaths, or sacrifices of protesters.
    \item Evokes empathy and moral outrage.
    \item \texttt{Eg:} \textit{"Elderly farmers spend nights on the streets for their rights.", "protesters continue to die, but their voices remain unheard."}
\end{itemize}

\item \textbf{Emphasizing Global and Celebrity Endorsements:} Articles highlighting international attention or celebrity support for protests to validate legitimacy.
\begin{itemize}[leftmargin=*, noitemsep, nolistsep]
    \item Cites solidarity tweets, global rallies, or NGO statements.
    \item Frames global support as moral validation.
    \item \texttt{Eg:} \textit{"Global icons from across continents have supported India’s farmers.", "The world is watching as India’s farmers fight for justice."}
\end{itemize}

\item \textbf{Accusing Media and Government of Manipulation:} Articles claiming media and government misrepresent or suppress protest realities.
\begin{itemize}[leftmargin=*, noitemsep, nolistsep]
    \item Mentions biased reporting or censorship.
    \item Suggests narrative control or propaganda.
    \item \texttt{Eg:} \textit{“The media is running government-fed lies about farmers’ violence.” “Independent journalists face harassment for showing ground truth.”}
\end{itemize}
\end{enumerate}

\paragraph{Pro-Govt Aligned Narratives (CAA):} 

\begin{enumerate}[leftmargin=*, noitemsep, nolistsep]
\item \textbf{Glorification of CAA:} Articles portraying the Citizenship Amendment Act as patriotic, humanitarian, or constitutionally sound.
\begin{itemize}[leftmargin=*, noitemsep, nolistsep]
    \item Emphasizes moral duty or compassion.
    \item Links CAA to India's civilizational ethos.
    \item \texttt{Eg:} \textit{"CAA offers refuge to persecuted minorities—it’s an act of compassion.", "This law strengthens India’s humanitarian spirit."}
\end{itemize}

\item \textbf{Glorification of the central government:} Articles presenting the government as protector of India’s sovereignty or integrity.
\begin{itemize}[leftmargin=*, noitemsep, nolistsep]
    \item Frames CAA/NRC as defense against illegal immigration.
    \item Appeals to national unity or security.
    \item \texttt{Eg:} \textit{"The government stands firm to protect India’s borders and identity."}
\end{itemize}



\item \textbf{Vilification of the Opposition:} Articles that morally discredit or delegitimize opposition parties, portraying them as politically motivated, anti-national, or obstructive to government initiatives.
\begin{itemize}[leftmargin=*, noitemsep, nolistsep]
\item Frames the opposition as divisive or acting against national interest.
\item Uses morally charged language to question credibility or intent.
\item \texttt{Eg:} \textit{"Opposition leaders are opposing CAA only to appease vote banks.", "They are obstructing reforms meant for the welfare of persecuted minorities."}
\end{itemize}

\item \textbf{Opposition Spreading Misinformation and Fear:} Articles that cognitively discredit the opposition or media by accusing them of spreading falsehoods, misinterpretations, or panic regarding the CAA/NRC.
\begin{itemize}[leftmargin=*, noitemsep, nolistsep]
\item Mentions deliberate fear-mongering or fake news campaigns.
\item Highlights ignorance or distortion of the law's provisions.
\item Frames unrest as a result of misinformation rather than a legitimate grievance.
\item \texttt{Eg:} \textit{"False propaganda is misleading citizens about CAA.", "Unfounded rumors are fueling unnecessary fear among minorities."}
\end{itemize}

\item \textbf{Anti-CAA protests as Pre-Planned and Funded:} Articles alleging protests are orchestrated or foreign-funded to defame India.
\begin{itemize}[leftmargin=*, noitemsep, nolistsep]
    \item Mentions NGOs, conspiracies, or hidden motives.
    \item Frames unrest as part of an international agenda.
    \item \texttt{Eg:} \textit{"The protests were planned months before the Act passed.", "Foreign agencies are behind the chaos."}
\end{itemize}

\item \textbf{Framing Anti-CAA protests as Subversive or Misguided:} Articles portraying protesters as misinformed, anti-national, or extremist.
\begin{itemize}[leftmargin=*, noitemsep, nolistsep]
    \item Frames protests as threats to national unity.
    \item Suggests participants are misled or ignorant.
    \item \texttt{Eg:} \textit{"protesters have no idea what the law even says."}
\end{itemize}

\item \textbf{Delegitimization of Critics:} Articles suggesting critics of CAA are biased or communal.
\begin{itemize}[leftmargin=*, noitemsep, nolistsep]
    \item Attacks left-wing commentators or state-aligned voices.
    \item Calls out intolerance disguised as patriotism.
    \item \texttt{Eg:} \textit{"Those protesters against CAA expose their deep-rooted prejudice."}
\end{itemize}
\end{enumerate}

\paragraph{Pro-Opp Aligned Narratives (CAA):} 

\begin{enumerate}[leftmargin=*, noitemsep, nolistsep]
\item \textbf{Vilification of CAA:} Articles framing the law as discriminatory, unconstitutional, or anti-minority.
\begin{itemize}[leftmargin=*, noitemsep, nolistsep]
    \item Highlights exclusion of Muslims or erosion of secularism.
    \item Uses moral or legal condemnation.
    \item \texttt{Eg:} \textit{"CAA legitimizes religious discrimination under the guise of compassion.", "This Act dismantles India’s secular foundation."}
\end{itemize}

\item \textbf{Vilification of the Central Government:} Articles accusing the government of authoritarianism, repression, or communal politics.
\begin{itemize}[leftmargin=*, noitemsep, nolistsep]
    \item Focuses on police brutality or suppression of dissent.
    \item Frames leadership as divisive or intolerant.
    \item \texttt{Eg:} \textit{"The government treats peaceful protesters as enemies."} 
\end{itemize}


\item \textbf{Glorifying Anti-CAA protesters:} Articles portraying protesters as brave defenders of democracy and justice.
\begin{itemize}[leftmargin=*, noitemsep, nolistsep]
    \item Describes protests as peaceful or youth-led.
    \item Frames protesters as voices of conscience.
    \item \texttt{Eg:} \textit{"Women of Shaheen Bagh became the conscience of the nation."} 
\end{itemize}

\item \textbf{Framing Anti-CAA protesters as Victims:} Articles emphasizing state violence or persecution against protesters.
\begin{itemize}[leftmargin=*, noitemsep, nolistsep]
    \item Focuses on casualties, injuries, or arrests.
    \item Highlights moral injustice or fear.
    \item \texttt{Eg:} \textit{"protesters face jail for exercising free speech."}
\end{itemize}
\end{enumerate}

\paragraph{Persuasive Technique Guidelines:} 

Persuasion techniques determine the common practice people use to propagate their agenda. In one of the very first work on propaganda analysis, \newcite{da-san-martino-etal-2019-fine} defined 18 unique persuasion techniques. We also adopt these in our work and extend them with two additional techniques: \textit{assertion} \cite{jowett2018propaganda}, presenting a statement as fact without supporting evidence, often to persuade through confidence rather than proof  and \textit{glittering generalities} (Virtue) \cite{bauer2024glittering} using emotionally appealing but vague and positive words to evoke approval without conveying specific meaning. We follow  the guidelines of \newcite{da-san-martino-etal-2019-fine} for persuasive technique annotation. 
.

\subsection{Annotation}

We employ nine annotators, aged between 20 - 40 (including five males and four females), for the annotation. All annotators are proficient in English and familiar with Indian politics. They represent a diverse background of research assistants, graduate/postgraduate students, and industrial researchers. In addition, two moderators oversee the annotation process and ensure quality control. All annotators/moderators are compensated according to the institute's norm.

\paragraph{Phase 1 -- Annotator Training: }
All annotators underwent a unified, multi-stage training process covering bias labeling, fine-grained narrative framing, and persuasive technique identification. 

\begin{itemize}[leftmargin=*, noitemsep, nolistsep]
    \item \textbf{Orientation Session:} Annotators first attended a structured orientation that outlined the objectives of the project. Each annotator received the detailed annotation manual, which described every category with definitions and examples. Trainers clarified distinctions such as factual criticism vs ideological bias and topic description vs narrative framing. This session ensured that all annotators shared a common conceptual foundation.
    
    
    \item \textbf{Pilot Annotation and Calibration: }Annotators independently labeled a  subset of articles spanning both events. After each round, we conducted calibration meetings in which all annotations were discussed and disagreements were resolved. Through these sessions, annotators refined shared interpretations and harmonized their application of the guidelines across all three levels. In each calibration cycle, IAA scores were computed for bias, narrative, and technique labels to quantify consistency and accuracy. We repeat the pilot and calibration step until stable annotator behavior is observed. 
\end{itemize}

\paragraph{Phase 2 -- Final Annotation:} The final annotation phase begins after achieving stable annotation consistency during training. At this stage, annotators annotate the articles independently. To assess the annotation quality, we compute inter-rater agreement on the annotation. For multi-class bias annotations, we compute Fleiss’ $\kappa$ \cite{fleiss1971measuring,carletta1996assessing} and obtain a score of $\kappa=0.611$. In multi-label fine-grained narrative annotations, we observe mean Fleiss’ $\kappa=0.605$ and $\kappa=0.613$ for Farmers' protest and CAA, respectively. We also compute mean pairwise Jaccard similarity \cite{jaccard1901etude} as 0.728 for Farmers' protests and 0.673 for CAA. For multi-label persuasive technique annotations. we obtain mean Fleiss’ $\kappa=0.312$ \& pairwise Jaccard similarity of 0.364.

\begin{table}[!t]
\centering
\setlength{\tabcolsep}{3pt} 
\resizebox{\columnwidth}{!}{
\begin{tabular}{l rrr rrr}
\toprule
\multirow{4}{*}{\textbf{Persuasive Techniques}} & \multicolumn{3}{c}{\textbf{CAA}} & \multicolumn{3}{c}{\textbf{FARMERS}} \\
\cmidrule(lr){2-4} \cmidrule(lr){5-7}
& \textbf{Train} & \textbf{Test} & \textbf{Total} & \textbf{Train} & \textbf{Test} & \textbf{Total} \\
\cmidrule(lr){2-4} \cmidrule(lr){5-7}
 & 537 & 131 & 668  & 480 & 118 & 598 \\

\midrule

\textbf{T1}: Causal Oversimplification & 51 & 12 & 63 & 30 & 4 & 34 \\
\textbf{T2}: Black-and-white Fallacy & 12 & 3 & 15 & 6 & 1 & 7 \\
\textbf{T3}: Straw Man & 30 & 7 & 37 & 6 & 4 & 10 \\
\textbf{T4}: Whataboutism & 23 & 6 & 29 & 3 & 2 & 5 \\ \textbf{T5}: Reductio ad Hitlerum & 11 & 3 & 14 & 2 & 1 & 3 \\
\textbf{T6}: Red Herring & 24 & 7 & 31 & 5 & 1 & 6 \\ \textbf{T7}: Loaded Language & 112 & 29 & 141 & 55 & 11 & 66 \\
\textbf{T8}: Name Calling and Labeling & 104 & 27 & 131 & 43 & 12 & 55 \\ 
\textbf{T9}: Appeal to Emotion, Fear, Prejudice & 144 & 36 & 180 & 70 & 17 & 87 \\
\textbf{T10}: Slogans & 64 & 12 & 76 & 27 & 8 & 35 \\ \textbf{T11}: Flag-Waving & 52 & 8 & 60 & 11 & 4 & 15 \\
\textbf{T12}: Exaggeration and Minimization & 21 & 5 & 26 & 20 & 5 & 25 \\
\textbf{T13}: Thought-Terminating Clichés & 46 & 11 & 57 & 10 & 2 & 12 \\
\textbf{T14}: Bandwagon & 19 & 4 & 23 & 13 & 2 & 15 \\
\textbf{T15}: Smears & 176 & 49 & 225 & 100 & 22 & 122 \\
\textbf{T16}: Obfuscation, intentional vagueness & 19 & 4 & 23 & 7 & 2 & 9 \\ 
\textbf{T17}: Doubt & 49 & 12 & 61 & 26 & 5 & 31 \\ \textbf{T18}: Appeal to Authority & 30 & 9 & 39 & 8 & 3 & 11 \\
\textbf{T19}: Assertion & 148 & 51 & 199 & 70 & 17 & 87 \\ \textbf{T20}: Glittering Generalities & 51 & 16 & 67 & 46 & 17 & 63 \\
\midrule
\textbf{Total} & \textbf{1186} & \textbf{311} & \textbf{1497} & \textbf{558} & \textbf{140} & \textbf{698} \\
\bottomrule

\end{tabular}}
\vspace{-3mm}
\caption{Statistics of Persuasive Techniques.}
\label{tab:stat_span}
\vspace{-5mm}
\end{table}

\subsection{Dataset statistics}
Our \dataset{} dataset comprises of 1,266 news articles, 668 in CAA and 598 in Farmers’ protest. Overall, 54\% of articles exhibit bias (Pro-Govt or Pro-Opp) in CAA, while the Farmers’ protest contains 30\% biased content. Ideology-wise narrative class distribution for both events is presented in Table \ref{tab:stat_narrative}. 
We note that observe right-leaning narratives dominate CAA, while Farmers’ protest is dominated by  left leaning narratives. In total, \dataset{} contains 1258 and 377 narrative labels in CAA and Farmers' protest events, respectively. 
Finally, we report statistics of persuasive techniques in Table \ref{tab:stat_span}. We observe that a few techniques (\textit{smears}, \textit{assertion}, \textit{appeal to emotion}, \textit{name calling}, \textit{loaded language}, etc.) dominate others in both events.

\begin{figure*}[!t]
\begin{center}
\includegraphics[width=\textwidth]{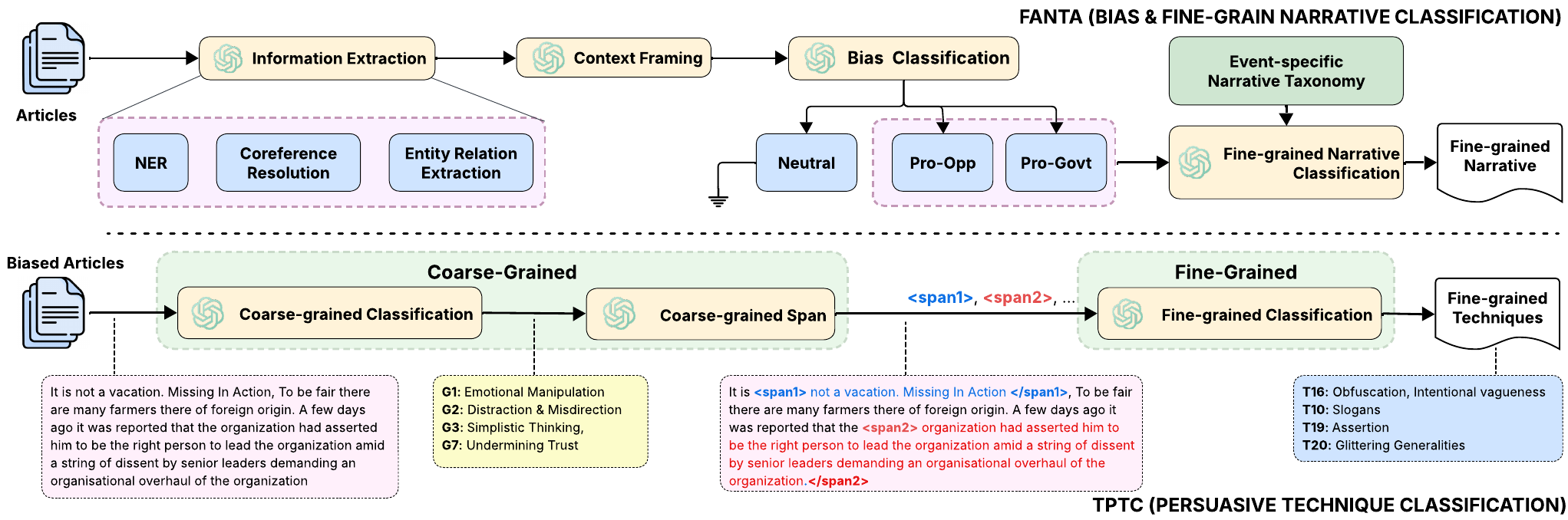}
\vspace{-5mm}
\caption{Overview of the proposed models architecture:
The upper pipeline illustrates \modelNarrative{} for the Bias and Fine-Grain Narrative Classification process. The lower pipeline illustrates \modelTechnique{} for the Persuasive Technique Classification process.}
\label{fig:model}
\end{center}
\vspace{-5mm}
\end{figure*}

\section{Proposed Methodology}

We propose two distinct novel frameworks, employing multi-hop prompting strategy, for the three classification tasks. The first framework, \modelNarrative{}, leverages the entity-relation dynamics in the article and its contextual framing for the article-bias classification. Subsequently, we employ an event-specific taxonomy for the fine-grained narrative classification.
The second framework, \modelTechnique{}, is a two-stage decomposition approach for the fine-grained persuasion techniques classification. It, at first, decomposes the article into identifying the conceptual category of persuasion, and subsequently, maps them into fine-grained persuasion techniques. A  high-level overview of the proposed frameworks is presented in Figure \ref{fig:model}.

\subsection{Bias and Narrative Classification}
We propose \modelNarrative{} an inference-only prompt-based classification model via multi-hop reasoning process \cite{wei2022chain}. Each hop incrementally aids the model's interpretation of the article, from factual foundation to ideological abstraction. 
We group these hops into two functional stages: \textbf{information extraction} and \textbf{context framing}. The outcomes of these two steps are utilized in the prompt for article-bias classification. Subsequently, for biased articles, we perform narrative classification following the event-specific taxonomy.

\paragraph{Information Extraction:} 
Entity-relation dynamics in an article is one of the foremost step in high-level reasoning comprehension. 
To achieve this, we begin with extracting structural information from each article in terms of NER, co-reference resolution, and entity-relation pair. Through NER and co-reference resolution, we identify all key entities as individuals, organizations, laws, political institutions, etc. along with their linking pronouns, aliases, and mentions and  mentioned in the article. Finally, we obtain relationship among entities through the entity-relation extraction module. By formalizing these relationships, we establish the necessary foundation for the analysis of the next hop.

\paragraph{Context Framing:}
Context framing refers to the process of determining how different actors, events, or institutions are positioned within the discourse. Building upon the fact that media bias rarely stems from isolated statements but from the interplay of actors/institutions relative to one another, we infer how these entities are framed within the same article. We hypothesize that bias is not always explicit in factual statements but it is often encoded in the tone, emphasis, and moral perspective adopted by the writer. By identifying framing patterns, we can capture how ideology manifests through language and thus, can infer the ideological alignment of an article’s portrayal. This step signifies the core novelty of \modelNarrative{} by providing interpretive foundation for bias reasoning, as depicted in Table \ref{tab:instance:framing}.

\paragraph{Article-Bias Classification:} Utilizing the contextual grounding, we perform a three-way bias classification for each article into \textit{pro-govt}, \textit{pro-opp}, or \textit{neutral} labels. This layered reasoning allows bias classification to function as an interpretive conclusion rather than a direct classification task.

\begin{table}[t]
    \centering
    \resizebox{\columnwidth}{!}{
    \begin{tabular}{p{5cm}p{3cm} p{6cm}}
    \toprule
         \bf Article & \bf Info Extraction & \bf Context Framing \\ \toprule
         NEW DELHI: <\textit{name}>, the leader of <\textit{party1}>, has criticized the central government for implementing the CAA, calling it unconstitutional. <\textit{name}> said, <\textit{party2}> has made unconstitutional law. It has been made based on religion, which goes against the right to equality, therefore I tore a copy of the bill in the Parliament. We were, are and continue to be against this." The <\textit{party1}> chief expressed optimism about his party's candidates performing well in the upcoming Lok Sabha elections. He clarified that <\textit{party1}> is not in alliance with any... & $\bullet$ <\textit{name}> is a leader of <\textit{party1}> and criticizes the CAA, indicating a direct opposition to <\textit{party2}> and its policies. \newline $\bullet$ <\textit{name}>’s remarks about the <\textit{party2}> abusing the Election Commission suggest a conflict between <\textit{party1}> and <\textit{party2}>, focusing on election integrity. $\bullet$ … & The article frames <\textit{name}> as a vocal opponent of the CAA and the policies of the <\textit{party2}>-led central government, emphasizing his legal and constitutional arguments against what he deems unconstitutional actions. Further, <\textit{name}> is depicted as actively campaigning for <\textit{party1}> and being involved in electoral strategies, signifying a proactive political approach. The mentioning of specific accusations against the <\textit{party2}> indicates a charged political atmosphere and reflects an ongoing rivalry. The article presents <\textit{name}> statements on electoral integrity and the political alliances within different states, maintaining a focus on <\textit{party1}> position in the political landscape. \\ \bottomrule
    \end{tabular}}
    \vspace{-3mm}
    \caption{An instance of information extraction and context framing in \modelNarrative{}.}
    \label{tab:instance:framing}
    \vspace{-3mm}
\end{table}

\begin{table*}[h!]
\centering

\resizebox{\textwidth}{!}{
\begin{tabular}{l ccc ccc ccc ccc ccc ccc}
\toprule

\multirow{3}{*}{\textbf{Prompt Settings}} & \multicolumn{9}{c}{\textbf{CAA}} & \multicolumn{9}{c}{\textbf{Farmer's Protest}} \\ \cmidrule(lr){2-10} \cmidrule(lr){11-19}
& \multicolumn{3}{c}{\textbf{Micro}} & \multicolumn{3}{c}{\textbf{Macro}} & \multicolumn{3}{c}{\textbf{Weighted}} & \multicolumn{3}{c}{\textbf{Micro}} & \multicolumn{3}{c}{\textbf{Macro}} & \multicolumn{3}{c}{\textbf{Weighted}} \\ \cmidrule(lr){2-4} \cmidrule(lr){5-7} \cmidrule(lr){8-10} \cmidrule(lr){11-13} \cmidrule(lr){14-16} \cmidrule(lr){17-19}

 & Pre & Rec & F1 & Pre & Rec & F1 & Pre & Rec & F1 & Pre & Rec & F1 & Pre & Rec & F1 & Pre & Rec & F1 \\ \midrule
RoBERTa  & 0.705 & 0.710 & 0.707 & 0.475 & 0.520 & 0.488 & 0.641 & 0.710 & 0.662 & 0.735 & 0.703 & 0.719 & 0.433 & 0.460 & 0.446 & 0.681 & 0.703 & 0.691  \\
BERT         & \textbf{0.766} & 0.626 & 0.689 & \textbf{0.733} & 0.580 & 0.641 & \textbf{0.777} & 0.626 & 0.685 & 0.738 & 0.670 & 0.702 & 0.433 & 0.435 & 0.434 & 0.683 & 0.670 & 0.676 \\
DistillBERT  & \textbf{0.766} & 0.626 & 0.689 & 0.532 & 0.457 & 0.475 & 0.715 & 0.626 & 0.645 & 0.761 & 0.729 & 0.745 & 0.462 & 0.428 & 0.432 & 0.678 & 0.729 & 0.691 \\
DeBERTa      & 0.713 & 0.664 & 0.688 & 0.520 & 0.483 & 0.471 & 0.697 & 0.664 & 0.640 & \textbf{0.771} & 0.712 & 0.740 & \textbf{0.750} & 0.642 & 0.658 & \textbf{0.831} & 0.712 & 0.749 \\ \midrule
GPT-4o-mini (Zero-Shot)   & 0.473 & 0.473 & 0.473 & 0.158 & 0.333 & 0.214 & 0.224 & 0.473 & 0.304 & 0.695 & 0.695 & 0.695 & 0.232 & 0.333 & 0.273 & 0.483 & 0.695 & 0.570 \\
GPT-4o-mini (Few-Shot)  & 0.374 & 0.374 & 0.374 & 0.194 & 0.344 & 0.245 & 0.231 & 0.374 & 0.285 & 0.542 & 0.542 & 0.542 & 0.248 & 0.278 & 0.262 & 0.477 & 0.542 & 0.507  \\

\midrule
\modelNarrative{} -- Mistral & 0.730 & 0.730 & 0.730 & 0.660 & 0.640 & 0.640 & 0.720 & 0.730 & 0.720 & \textbf{0.771} & 0.771 & \textbf{0.771} & 0.743 & 0.654 & 0.652 & 0.793 & 0.771 & 0.751 \\
\modelNarrative{} -- Gemma  & 0.530 & 0.530 & 0.530 & 0.630 & 0.530 & 0.420 & 0.730 & 0.530 & 0.420 & 0.474 &	0.474 &	0.474 &	0.531 &	0.704 &	0.467 &	0.759 &	0.474 &	0.475 \\

\modelNarrative{} -- GPT-4o-mini & 0.670 & \textbf{0.830} & \textbf{0.740} & 0.590 & \textbf{0.790} & \textbf{0.660} & 0.700 & \textbf{0.830} & \textbf{0.740} & 0.730 & \textbf{0.810} & 0.770 & 0.650 & \textbf{0.830} & \textbf{0.700} & 0.780 & \textbf{0.810} & \textbf{0.780} \\
\modelNarrative{} -- GPT-4o-mini (Concise) & 0.520 & 0.420 & 0.410 & 0.520 & 0.420 & 0.410 & 0.520 & 0.520 & 0.480 & 0.720 & 0.720 & 0.720 & 0.570 & 0.490 & 0.500 & 0.700 & 0.710 & 0.680 \\

\bottomrule
\end{tabular}
}
\caption{Article-Bias Classification: Comparative Results}
\label{tab:Bias_Model}
\end{table*}

\paragraph{Narrative Classification:} Following the article-bias classification, we proceed to fine-grained narrative classification for biased articles -- i.e., article identified as non-neutral in the previous stage. We utilize the biased label (pro-govt/pro-opp) to limit the underlying narrative classes in the taxonomy hierarchy for our input prompts in fine-grained narrative classification. The predicted narrative class, together with the context framing, explains the underlying storyline of the biased article with appropriate justification in term of contextual reasoning.

\subsection{Persuasive Technique Classification}

Prior research has demonstrated that LLMs, despite their strong linguistic reasoning capabilities, struggle to accurately detect persuasive techniques in zero- or few-shot settings \cite{jose2025large, hasanain-etal-2024-gpt}. We argue that persuasion is a multi-layered communicative phenomenon and often realized through subtle contextual cues rather than being explicit in nature. As a result, identifying persuasive technique in a text becomes challenging in a single inference step for LLMs. To address this challenge, we propose \modelTechnique{} that decomposes the problem into coarse-grained and fine-grained analysis. This decomposition allows the model to progressively reason about the conceptual category of persuasion at the coarse-level prior to mapping it fine-level techniques. Such a design mimics the natural cognitive process of human annotators, who first recognize the general intent and then identify the precise persuasion device.

\begin{table}[!h]
\centering

\resizebox{\columnwidth}{!}{
\begin{tabular}{l p{5cm} p{8cm}}
\toprule
& \bf Coarse-Grain & \bf Fine-Grain\\ \midrule
G1 & Emotional Manipulation & Appeal to emotion/fear/prejudice; Loaded Language; Flag-Waving; Name Calling and Labeling; Glittering Generalities; Slogans. \\ 
G2 & Distraction and Misdirection & Red Herring; Whataboutism; Obfuscation; Intentional Vagueness; Straw Man. \\ 
G3 & Simplistic Thinking & Causal Oversimplification; Black-and-white Fallacy; Thought-Terminating Clichés; Assertion. \\ 
G4 & Attack and Defamation &  Name Calling and Labeling; Smears; Reductio ad Hitlerum. \\ 
G5 & Manipulation through Popularity & Appeal to Authority; Bandwagon. \\ 
G6 & Misrepresentation and Distortion & Straw Man; Exaggeration and Minimization. \\ 
G7 & Undermining Trust &  Doubt; Obfuscation; Intentional Vagueness. \\
\bottomrule
\end{tabular}}

\vspace{-5mm}
\end{table}

In the first stage, we organize the existing 20 fine-grained persuasive techniques into seven coarse-grained categories based on their abstract-level relatedness and functional similarity in terms of persuasion. 
Each category represents a shared communicative goal and serves as a conceptual scaffold that enables the model to establish a broader understanding of persuasion across fine-grained techniques. 
Utilizing this framework, at first, we extract the coarse-level persuasive categories and their collective textual grounding as spans. Subsequently, we utilize the category labels to obtain the subset of fine-grained techniques belonging to that category to construct targeted prompts and spans to locate the potential regions-of-interest. We argue that by limiting the contextual region-of-interests, we allow the model to analyze the persuasive tone at a much finer-level for the technique identification.

\begin{table*}[h!]
\centering

\resizebox{\textwidth}{!}{
\begin{tabular}{l ccc ccc ccc ccc ccc ccc}
\toprule

\multirow{3}{*}{\textbf{Prompt Settings}} & \multicolumn{9}{c}{\textbf{CAA}} & \multicolumn{9}{c}{\textbf{Farmer's Protest}} \\ \cmidrule(lr){2-10} \cmidrule(lr){11-19}
& \multicolumn{3}{c}{\textbf{Micro}} & \multicolumn{3}{c}{\textbf{Macro}} & \multicolumn{3}{c}{\textbf{Weighted}} & \multicolumn{3}{c}{\textbf{Micro}} & \multicolumn{3}{c}{\textbf{Macro}} & \multicolumn{3}{c}{\textbf{Weighted}} \\ \cmidrule(lr){2-4} \cmidrule(lr){5-7} \cmidrule(lr){8-10}  \cmidrule(lr){11-13} \cmidrule(lr){14-16} \cmidrule(lr){17-19}

 & Pre & Rec & F1 & Pre & Rec & F1 & Pre & Rec & F1 & Pre & Rec & F1 & Pre & Rec & F1 & Pre & Rec & F1 \\ \midrule
RoBERTa    & \textbf{0.744} & 0.379 & 0.502 & 0.518 & 0.234 & 0.299 & 0.708 & 0.379 & 0.457 & 0.714 & 0.671 & 0.692 & 0.410 & 0.343 & 0.345 & 0.726 & 0.671 & 0.684 \\
BERT      & 0.646 & 0.399 & 0.493 & 0.502 & 0.289 & 0.337 & 0.618 & 0.399 & 0.448 & 0.712 & 0.664 & 0.688 & 0.183 & 0.226 & 0.201 & 0.604 & 0.664 & 0.629 \\
DistillBERT  & 0.664 & 0.497 & 0.568 & 0.542 & 0.366 & 0.405 & 0.678 & 0.497 & 0.555 & 0.621 & 0.584 & 0.602 & 0.163 & 0.212 & 0.179 & 0.562 & 0.584 & 0.561 \\
DeBERTa   & 0.680 & 0.444 & 0.538 & 0.350 & 0.251 & 0.269 & 0.566 & 0.444 & 0.471 & 0.681 & 0.658 & 0.669 & 0.211 & 0.236 & 0.221 & 0.621 & 0.658 & 0.637 \\ \midrule
GPT-4o-mini (Zero-Shot)  & 0.340 & 0.200 & 0.250 & 0.190 & 0.120 & 0.090 & 0.300 & 0.200 & 0.150 & 0.300 & 0.450 & 0.367 & 0.119 & 0.279 & 0.130 & 0.470 & 0.460 & 0.420 \\
GPT-4o-mini (Few-Shot)    & 0.060 & 0.100 & 0.080 & 0.020 & 0.330 & 0.040 & 0.007 & 0.100 & 0.014 & 0.100 & 0.240 & 0.150 & 0.048 & 0.298 & 0.077 & 0.050 & 0.235 & 0.082 \\ \midrule
\modelNarrative{} -- Mistral   & 0.650 & 0.520 & 0.580 & 0.580 & 0.410 & 0.430 & 0.700 & 0.520 & 0.550 & \textbf{0.757} & 0.691 & 0.723 & 0.522 & 0.381 & 0.427 & 0.746 & 0.691 & 0.696 \\
\modelNarrative{} -- Gemma   & 0.400 & 0.420 & 0.410 & 0.450 & 0.380 & 0.320 & 0.590 & 0.420 & 0.370 & 0.370 & 0.563 & 0.446 & 0.336 & \textbf{0.792} & 0.393 & 0.676 & 0.563 & 0.482 \\
\modelNarrative{} -- GPT-4o-mini & 0.654 & 0.810 & 0.724 & 0.564 & 0.768 & 0.632 & \textbf{0.708} & 0.810 & 0.732 & 0.732 & 0.805 & 0.767 & \textbf{0.702} & 0.780 & \textbf{0.709} & 0.775 & 0.805 & 0.772 \\

\modelNarrative{} -- GPT-4o-mini (Concise) & 0.670 & \textbf{0.830} & \textbf{0.740} & \textbf{0.590} & \textbf{0.790} & \textbf{0.660} & 0.700 & \textbf{0.830} & \textbf{0.740} & 0.730 & \textbf{0.810} & \textbf{0.770} & 0.650 & 0.830 & 0.700 & \textbf{0.780} & \textbf{0.810} & \textbf{0.780} \\

\bottomrule
\end{tabular}
}
\vspace{-3mm}
\caption{Fine-Grained Narrative Classification: Comparative Results.}
\label{tab:results_narrative}
\end{table*}

\begin{table*}[htbp]

\centering

\resizebox{\textwidth}{!}{

\begin{tabular}{l ccc ccc ccc ccc ccc ccc}

\toprule

\multirow{3}{*}{\textbf{Prompt Settings}} & \multicolumn{9}{c}{\textbf{CAA}} & \multicolumn{9}{c}{\textbf{Farmer's Protest}} \\ \cmidrule(lr){2-10} \cmidrule(lr){11-19}

& \multicolumn{3}{c}{\textbf{Micro}} & \multicolumn{3}{c}{\textbf{Macro}} & \multicolumn{3}{c}{\textbf{Weighted}} & \multicolumn{3}{c}{\textbf{Micro}} & \multicolumn{3}{c}{\textbf{Macro}} & \multicolumn{3}{c}{\textbf{Weighted}} \\ \cmidrule(lr){2-4} \cmidrule(lr){5-7} \cmidrule(lr){8-10}  \cmidrule(lr){11-13} \cmidrule(lr){14-16} \cmidrule(lr){17-19}

 & Pre & Rec & F1 & Pre & Rec & F1 & Pre & Rec & F1 & Pre & Rec & F1 & Pre & Rec & F1 & Pre & Rec & F1 \\




\midrule

RoBERTa   & \textbf{0.749} & 0.329 & 0.457 & 0.206 & 0.118 & 0.140 & 0.509 & 0.329 & 0.376 & 0.687 & 0.509 & 0.585 & 0.156 & 0.167 & 0.157 & 0.495 & 0.509 & 0.496 \\
BERT      & 0.689 & 0.364 & 0.476 & \textbf{0.294} & 0.166 & 0.201 & 0.545 & 0.364 & 0.423 & 0.719 & 0.455 & 0.557 & 0.156 & 0.130 & 0.136 & 0.522 & 0.455 & 0.476 \\
DistillBERT & 0.699 & 0.375 & 0.488 & 0.221 & 0.157 & 0.181 & 0.502 & 0.375 & 0.423 & 0.669 & 0.405 & 0.504 & 0.134 & 0.119 & 0.124 & 0.491 & 0.405 & 0.439 \\
DeBERTa     & 0.730 & 0.342 & 0.466 & 0.235 & 0.134 & 0.162 & 0.520 & 0.342 & 0.394 & 0.674 & 0.423 & 0.520 & 0.201 & 0.128 & 0.125 & 0.568 & 0.423 & 0.439 \\ \midrule
GPT-4o-mini (Zero-Shot) & 0.208 & 0.517 & 0.297 & 0.167 & 0.385 & 0.190 & 0.272 & 0.517 & 0.298 & 0.098 & 0.398 & 0.157 & 0.147 & 0.342 & 0.106 & 0.293 & 0.398 & 0.159 \\
GPT-4o-mini (Few-Shot) & 0.210 & 0.500 & 0.300 & 0.198 & \textbf{0.390} & 0.200 & 0.270 & 0.500 & 0.297 & 0.090 & 0.345 & 0.143 & 0.739 & 0.300 & 0.084 & 0.129 & 0.345 & 0.134 \\ \midrule


\modelTechnique{} -- Mistral & 0.250 & 0.292 & 0.269 & 0.246 & 0.236 & 0.201 & 0.451 & 0.292 & 0.301 & 0.188 & 0.252 & 0.215 & 0.202 & 0.256 & 0.171 & 0.487 & 0.252 & 0.266 \\
\modelTechnique{} -- Gemma & 0.440 & 0.270 & 0.330 & 0.259 & 0.169 & 0.174 & 0.401 & 0.268 & 0.283 & 0.600 & 0.418 & 0.493 & 0.294 & 0.191 & 0.208 & 0.513 & 0.418 & 0.421 \\
\modelTechnique{} -- GPT-4o-mini & 0.728 & \textbf{0.728} & \textbf{0.728} & 0.219 & 0.234 & \textbf{0.214} & \textbf{0.840} & \textbf{0.766} & \textbf{0.784} & \textbf{0.847} & \textbf{0.847} & \textbf{0.847} & \textbf{0.814} & \textbf{0.887} & \textbf{0.834} & \textbf{0.820} & \textbf{0.910} & \textbf{0.853} \\
\bottomrule

\end{tabular}}
\vspace{-3mm}
\caption{Fine-Grained Persuasive Technique Classification: Comparative Results. }
\label{tab:PersuasiveTechnique_classification_Final}
\end{table*}

\begin{table*}[!ht]

\centering

\resizebox{\textwidth}{!}{

\begin{tabular}{ p{2.2cm} p{6.5cm} l p{15cm}}

\toprule

\textbf{Task} & \textbf{Ground truth} & \textbf{Model} & \textbf{Prediction} \\ \midrule


\multirow[t]{6}{2.2cm}{Persuasive Technique Classification} & \multirow[t]{5}{6.5cm}{[appeal to emotion/fear/prejudice; assertion; glittering generalities (virtue); loaded language; slogans; smears;]} & GPT-4o-mini (FS) & \textcolor{darkgreen}{[Assertion;]} \textcolor{red}{[Causal Oversimplification; Exaggeration and Minimization; Black-and-white Fallacy; Name Calling and Labeling; Appeal to Authority; Flag-Waving;} \textcolor{blue}{appeal to emotion/fear/prejudice; glittering generalities (virtue); loaded language; slogans; smears;]} \\ \cmidrule(lr){3-4}



& & \modelTechnique{} -- GPT-4o-mini & \textcolor{darkgreen}{[appeal to authority; smears;} \textcolor{red}{bandwagon; exaggeration and minimization; name calling and labeling; reductio ad hitlerum; straw man;} \textcolor{blue}{assertion; glittering generalities (virtue); loaded language; slogans;]} \\ \midrule


\multirow[t]{6}{2.2cm}{Narrative Classification} & \multirow[t]{5}{6.5cm}{[glorification of CAA; vilification of the opposition; opposition spreading misinformation and fear; delegitimization of critics; glorification of the central government;]} & GPT-4o-mini (FS) & \textcolor{red}{[Vilification of CAA; Vilification of Central Government; Glorifying Anti-CAA Protesters; Framing Anti-CAA Protesters as Victims;} \textcolor{blue}{Glorification of CAA; Vilification of the opposition; opposition spreading misinformation and fear; Delegitimization of critics; Glorification of the central government;]} \\ \cmidrule(lr){3-4}



& & \modelNarrative{} -- GPT-4o-mini & [\textcolor{darkgreen}{Vilification of the opposition; opposition spreading misinformation and fear; Glorification of the central government;} \textcolor{blue}{Glorification of CAA; Delegitimization of critics;]} \\

\bottomrule

\end{tabular}
}
\vspace{-3mm}
\caption{Error analysis: \textcolor{darkgreen}{True Positives}, \textcolor{red}{False Positives}, \textcolor{blue}{False Negatives}.}
\label{tab:error}
\vspace{-3mm}
\end{table*}

\section{Experiments and Results}

\paragraph{Baselines and Comparative Systems:} Existing works \cite{da-san-martino-etal-2020-semeval,dimitrov-etal-2024-semeval,piskorski-etal-2025-semeval,dimitrov-etal-2021-semeval} have relied heavily on the fine-tuned PLMs due to their superior performance on propaganda-related tasks. We also comply and utilize several fine-tuned PLMs (BERT \cite{devlin2019bert}, RoBERTa \cite{liu2019roberta}, DistillBERT \cite{sanh2019distilbert}, and DeBERTa \cite{he2020deberta}) for comparison. In addition, we also explore LLMs paradigms with GPT-4o-mini in zero- and few-shot settings as well. Moreover, we also ablate our proposed frameworks (i.e, \modelNarrative{} and \modelTechnique{}) with Gemma-3-12b \cite{gemma_2025} and Mistral-8x7B-v0.1\cite{mixtral_8x7b}. Further, for the article-bias and narrative classification, we try a concise variant, \modelNarrative{} (Concise), where we perform NER, Co-reference Resolution and entity-relation pair extraction in a single-hop.

\paragraph{Article-Bias Classification:} We report results of article-bias classification in Table \ref{tab:Bias_Model}. We observe that, in most of the cases, PLMs report better precision scores than LLMs based systems. Furthermore, GPT-4o-mini in zero- and few-shot settings yield inferior results against all competition systems. This is in-line with existing works which showed poor adaptability of LLMs in propaganda settings. However, on the contrary, leveraging explicit information extraction and context framing, our proposed framework, \modelNarrative{}, outperform PLMs on recall and F1-score across both events.  

\paragraph{Narrative Classification:} We observe similar trends in the narrative classification step as well in Table \ref{tab:results_narrative}. Moreover, the concise framework further assists the model in improved prediction. This indicates that explicitly merging entity and relation information strengthens discourse-level coherence by enabling the model to reason about who interacts with whom within ideological frames. Overall, the structured reasoning approach integrating entity-relational analysis and context framing consistently outperforms all fine-tuned baselines and prompting variants across all metrics.

\paragraph{Persuasive Technique Identification:} Table \ref{tab:PersuasiveTechnique_classification_Final} presents the comparative results for the fine-grained persuasive technique classification for both events. 
We observe moderate performance of PLMs -- in the range of [0.457, 0.488] micro-F1, [0.140, 0.201] macro-F1, and [0.376, 0.423] weighted-F1 on the CAA dataset and in the range of [0.504, 0.585] micro-F1, [0.124, 0.157] macro-F1, and [0.439, 0.496] weighted-F1 on the Farmers' dataset. In contrast, our \modelTechnique{} achieves a substantial leap in performance, attaining 0.728 micro-F1, 0.214 macro-F1, and 0.784 weighted-F1 on CAA. Similarly, \modelTechnique{} yields 0.847 micro-F1, 0.834 macro-F1, and 0.853 weighted-F1 on Farmers’ protest. We argue that the performance-gain is due to our two-stage hierarchical prompting design, which recognizes the coarse persuasive intent before mapping it to fine-grained persuasive techniques.

\begin{wrapfigure}{r}{0.4\columnwidth}
\centering
\includegraphics[width=0.35\columnwidth]{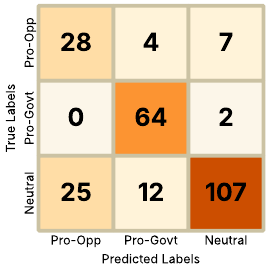}
\caption{Confusion matrix for Article-Bias Classification.}
\label{fig:Bias_CM}
\end{wrapfigure}
\paragraph{Error Analysis:}

We report our error analyses in Figure \ref{fig:Bias_CM} and Table \ref{tab:error}. For the bias classification task, the confusion matrix in Figure \ref{fig:Bias_CM} shows that the model performs well overall, particularly in identifying right-leaning articles, with a high precision of 64 \textit{true-positives} and only 2 \textit{false-negatives}. The main source of error stems from the model's difficulty in distinguishing neutral articles from biased ones. This suggests that the model is highly sensitive to political keywords, sometimes over interpreting factual reporting as ideologically biased.

On the other hand, a qualitative analysis of the model's predictions reveals challenges in fine-grained classification for both narrative and technique classification tasks. In most of the cases, we observe that models make partially correct predictions, capturing some narrative themes while missing the complete picture. However, we also notice that proposed model usually predict within the ideological polarity.

\section{Conclusion}

In this paper, we proposed a novel fine-grained narrative classification for biased articles to unveil the storyline at play. To support this, we developed the first ideologically grounded fine-grained narrative dataset for Indian news media, covering two recent socio-political events (CAA and the Farmers’ protest) and annotated them across three levels: bias, narrative, and persuasive techniques. Our proposed multi-stage hierarchical models, \modelNarrative{} and \modelTechnique{} leveraged contextual reasoning over entities and their relations to uncover how ideological bias manifests through structured narratives and rhetorical persuasion. We observed that, LLMs, when provided with decomposed persuasive intent adapts efficiently for the fine-grained classifications for both narrative and persuasive techniques. Overall, these findings validate that hierarchical reasoning enables LLMs to bridge micro-level persuasive cues and macro-level ideological framing, capturing propaganda as a multi-layered communicative phenomenon. We believe this work lays the groundwork for narrative-aware propaganda analysis in diverse, multilingual contexts, and provides a strong empirical benchmark for future research on media framing, ideology, and persuasion.

\section{Ethical Considerations}

The \dataset{} dataset was created with careful attention to ethical issues related to data collection, annotation, and release. All articles included in the dataset were collected from publicly accessible news outlets, and no personally identifiable information (PII) beyond what was already published by the media was retained. The dataset does not include user-generated content such as social media posts, thereby avoiding concerns around individual consent or inadvertent disclosure of private data.

All news outlets were selected based on their rating on the Media Bias/Fact Check (MBFC)\footref{fn:mbfc} platform to ensure a balanced representation of ideological perspectives. We emphasize that we do not glorify, diminish, support, or oppose any of these outlets, they are included solely for research purposes to study media bias, framing, and propaganda within a controlled, transparent framework.

Nevertheless, the dataset reflects the biases, inaccuracies, and manipulative strategies present in the sources. Some outlets were deliberately chosen to include those flagged for poor sourcing or failed fact checks, some articles may contain offensive language, or extreme ideological framing. These elements are integral to studying propaganda but may reproduce harmful discourse if taken out of context. To mitigate this risk, the dataset is intended strictly for academic and research purposes. Users are encouraged to apply critical thinking when interpreting model outputs trained on the data. 

Another consideration arises from the subjectivity of annotations. Persuasive techniques and narrative frames are inherently interpretive categories, and annotators may bring their own perspectives into the labeling process. While we employed multiple rounds of training, pilot annotations, and consensus building to reduce subjectivity, some level of disagreement is unavoidable. Researchers using the dataset should therefore treat annotations as a curated but fallible representation rather than the absolute truth. 

From a border perspective, there is a risk that models trained on \dataset{} could be misused for purposes such as automated censorship, political profiling, or targeted manipulation of readers. We strongly caution against such applications. The dataset is released to advance transparency in media analysis, to support work in computational social science and NLP, and to foster a deeper understanding of how propaganda functions in public discourse. 

Finally, we recognize that the dataset focuses on two politically sensitive events in India. Our goal is not to endorse or delegitimize any political position but to provide a resource for studying persuasion in context. By making the data openly available, we hope to support a diversity of perspectives and to enable critical research on the media ecosystem in the Global South, which remains underrepresented in computational resources.

\section{Limitations}

While \dataset{} dataset advances the study of propaganda and ideological narratives in Indian media, several limitations remain. The dataset focuses on two major socio-political events (CAA/NRC and the Farmers' Protest), which may not capture the full diversity of propaganda strategies across other domains or time periods. Our models, despite leveraging multi-hop reasoning, rely primarily on textual cues and do not incorporate multimodal evidence such as images or metadata. Furthermore, the experiments are limited to English news sources, extending the framework to multilingual or code-mixed Indian media would further enhance its generalizability. Future work will address these limitations by expanding event coverage, integrating multimodal reasoning, and incorporating multilinguality.

\section{Bibliographical References}\label{sec:reference}
\bibliographystyle{lrec2026-natbib}
\bibliography{lrec2026}


\end{document}